\documentclass[letterpaper, 10 pt, conference]{ieeeconf}  

\makeatletter
\def\endthebibliography{%
	\def\@noitemerr{\@latex@warning{Empty `thebibliography' environment}}%
	\endlist
}
\makeatother

\IEEEoverridecommandlockouts                              

\usepackage{tikz}
\usepackage{textcomp}
\usepackage[normalem]{ulem}
\usepackage{graphicx}
\usepackage{xcolor}
\definecolor{shadecolor}{rgb}{1,.8,.3}
\usepackage{algorithm}
\usepackage[noend]{algpseudocode}
\usepackage{varwidth}
\usepackage{amsmath,amssymb}

\title{\LARGE \bf
Toward Synergic Learning for Autonomous Manipulation of Deformable Tissues via Surgical Robots: An Approximate Q-Learning Approach
}

\author{Sahba Aghajani Pedram$^{\dagger, 1}$, Peter Walker Ferguson$^{\dagger,1}$, Changyeob Shin$^{1}$, \\ Ankur Mehta$^{2}$, Erik P. Dutson$^{3}$, Farshid Alambeigi$^{4}$, and Jacob Rosen$^{1}$
\thanks{$\dagger$ Sahba Aghajani Pedram and Peter Walker Ferguson contributed equally to this work.}
\thanks{*Research supported by National Science Foundation award IIS-1227184: Multilateral Manipulation by Human-Robot.}
\thanks{$^{1}$Sahba Aghajani Pedram, Peter Walker Ferguson, Changyeob Shin, and Jacob Rosen are with the Mechanical and Aerospace Engineering Department, University of California at Los Angeles (UCLA), Los Angeles, CA, USA
{\tt\small\{sahbaap,pwferguson,shinhujune,jacobrosen\}
@ucla.edu}}%
\thanks{$^{2}$ Ankur Mehta is with the Electrical and Computer Engineering Department, University of California, Los Angeles (UCLA), CA, USA
        {\tt\small mehtank@ucla.edu}}%
\thanks{$^{3}$Erik Dutson is with the Department of Surgery, David Geffen School of Medicine, University of California at Los Angeles (UCLA), Los Angeles, CA 90095, USA 
        {\tt\small EDutson@mednet.ucla.edu}}%
\thanks{Farshid Alambeigi is  with the Department of Mechanical Engineering,  University of Texas at Austin, Austin, TX, USA, 78712. {\tt\footnotesize farshid.alambeigi@austin.utexas.edu}}
}

\begin{document}

\maketitle
\thispagestyle{empty}
\pagestyle{empty}

\begin{abstract}
In this paper, we present a synergic learning algorithm to address the task of indirect manipulation of an unknown deformable tissue. Tissue manipulation is a common yet challenging task in various surgical interventions, which makes it a good candidate for robotic automation. We propose using a linear approximate Q-learning method in which  human knowledge contributes to selecting useful yet simple features of tissue manipulation while the algorithm learns to take optimal actions and accomplish the task. The algorithm is implemented and evaluated on a simulation using the OpenCV and CHAI3D libraries. Successful simulation results for four different configurations which are based on realistic tissue manipulation scenarios are presented. Results indicate that with a careful selection of relatively simple and intuitive features, the developed Q-learning algorithm can successfully learn an optimal policy without any prior knowledge of tissue dynamics or camera intrinsic/extrinsic calibration parameters.
\end{abstract}

\section{Introduction}
Robot-Assisted Surgery (RAS) is becoming the norm  of many operating room procedures, as it enables enhanced precision, dexterity, and feedback. Compared to conventional minimally invasive surgery (MIS) and/or open surgery, however, RAS restricts paramount information such as haptic feedback and direct vision from the surgeons. These deficiencies make surgical sub-tasks such as suturing or tissue manipulation very challenging for, and demand high cognitive loads from, surgeons \cite{kapoor2005spatial}. To reduce the surgeon's burden, recent work has begun on automation of specific MIS tasks including suturing \cite{pedram2017autonomous, sen2016automating, shademan2016supervised}, tissue manipulation \cite{alambeigi2018toward,alambeigi2019autonomous}, tissue dissection \cite{murali2015learning}, and drilling \cite{coulson2008autonomous, alambeigi2019use}. Robotic automation of these tasks would place surgeons in a supervisory position and could reduce the physical and cognitional strain of the operation. Hence, robotic automation has great potential to improve surgery, and therefore patients' outcome.

\begin{figure}[t!]
       \centering
       \framebox{\parbox{3.3in}{\includegraphics[scale=0.165,trim={1.3cm 0.4cm 2.5cm 1.5cm},clip]{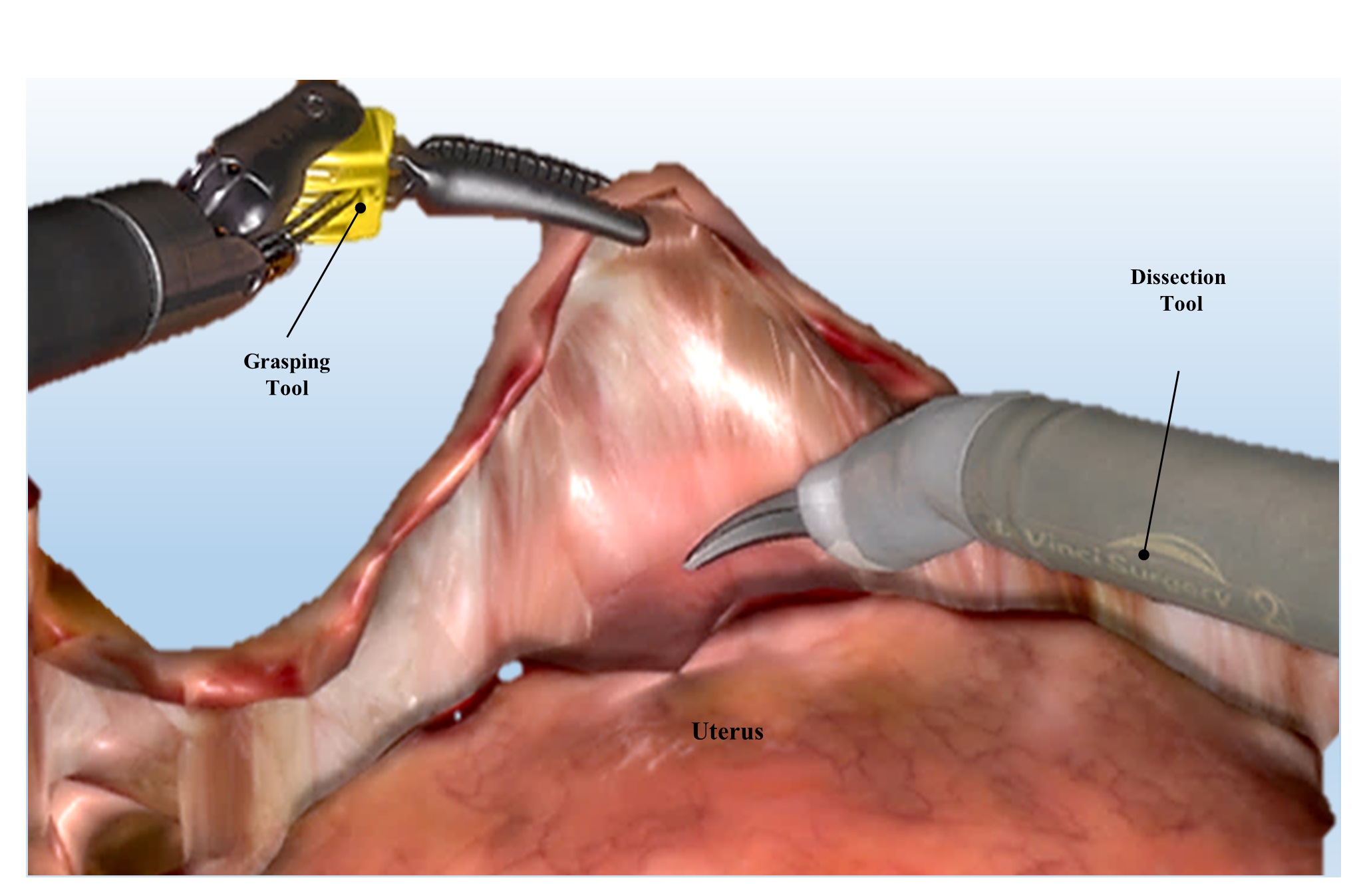}}}
      \caption{Example of an MIS tissue manipulation during a tissue debridement procedure.}
       \label{fig:SoftTissue}
 \end{figure}

A particular task of interest in various MIS interventions (e.g. tumor dissection and tissue debridement) is tissue manipulation, as shown in Fig. \ref{fig:SoftTissue}. Due to the uncertain deformation behavior of tissue, however, this  manipulation task is challenging even for expert clinicians. The main source of this complication are homogeneous/heterogeneous  physical or geometrical properties of deformable tissue such as stiffness and viscoelasticity which are unknown and hard-to-model \cite{alambeigi2018robust}. Recent advances in machine/robot learning fields have enabled them to learn very complicated mappings and/or tasks  \cite{krizhevsky2012imagenet,levine2016end}. 
Hence, in order to lessen the surgeon's burden during RAS, learning-based autonomous tissue manipulation strategies are reasonable paths to explore.


 \begin{figure*}[t!]
    \centering
    \includegraphics[scale=0.19, trim={0cm 0cm 0cm 0cm},clip]{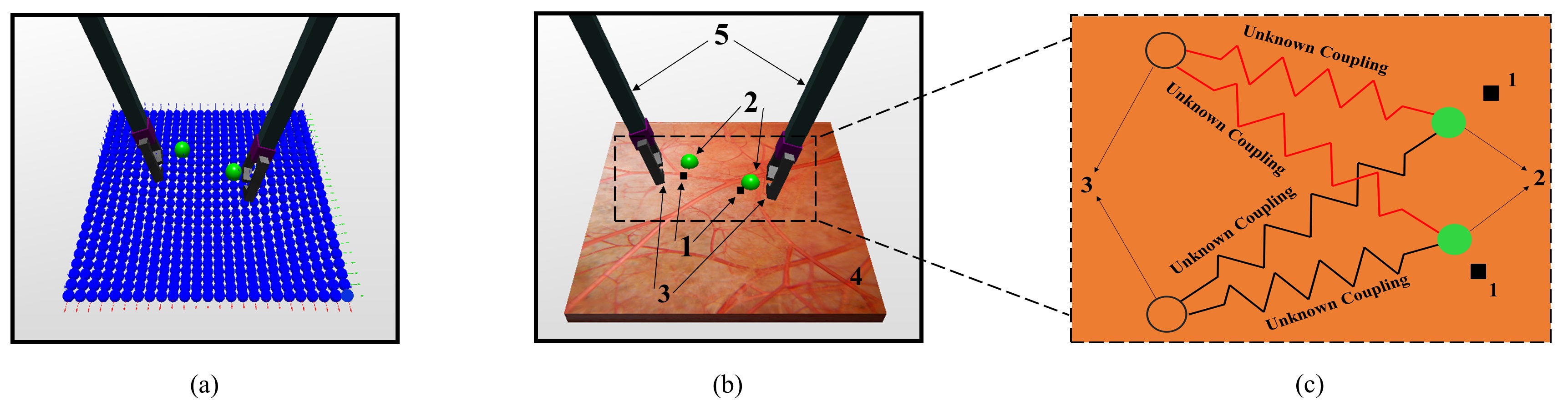}
    \caption{Simulation Environment. (a) All nodes used in the gel simulation (blue) with TTP highlighted (green). (b) Components of interest: 1) IDPs, 2) TTPs, 3) TGPs, 4) simulated tissue, and 5) graspers. (c) Representation of intuitive coupling between TTPs and actions of TGPs (1, 2, and 3 are as defined).}
    \label{SimEnvironment}
\end{figure*}

Several researchers have attempted different methods to address the task of deformable object/tissue manipulation. These methods can be classified into two broad categories based on whether they deploy human knowledge about physics of the problem or purely rely on the machine learning capabilities. The first category, which we refer to as \textit{human-knowledge driven} algorithms, incorporate explicit human knowledge about the model of the tissue and/or manipulation task. For example, indirect positioning of a deformable object based on a linearized second order model was proposed in \cite{hirai2000indirect}.
Reference \cite{jimenez2012survey} comprehensively reviews manipulation of deformable objects where algorithms deployed human knowledge in their methodologies explicitly. The second category of methods, which we refer to as \textit{machine-knowledge driven} or \textit{data driven} algorithms, encompass algorithms that  machines deploy data to build knowledge of the tissue and/or manipulation task. For example, an optimization framework with a an online Jacobian update was used to learn the dynamics of the tissue manipulation task \cite{alambeigi2018toward}.

Due to unknown or large
inter- and intra-subject variability between biological tissues in complex surgical scenarios, data driven approaches may be viewed as a more appropriate solution. However, these methods do not incorporate prior human knowledge into their solution, and thus may require large amounts of data to train which is both expensive and time-consuming \cite{bishop2006pattern}. For real-world robotics applications, methods of choice are data-driven algorithms which use more suitable and task-specific representations \cite{peters2010relative,deisenroth2013survey,gu2016continuous}. Hence, in the case of robotic tissue manipulation, a synergic learning (SL) algorithm which combines basic human knowledge about the physics of the problem with a learning algorithm is of interest.

To explore the performance of a SL algorithm in autonomous manipulation of deformable tissues, we propose an approximate Q-learning algorithm. For this algorithm, human knowledge contributes to selecting useful yet simple features of tissue manipulation while the algorithm learns to take optimal actions and accomplish the task. Q-learning \cite{watkins1992q,watkins1989learning} is among the most popular reinforcement learning algorithms in which the agent learns to take optimal actions without explicitly building knowledge of the system dynamics and/or control. Q-learning has already been successfully applied to a range of challenging robotics problems \cite{kober2013reinforcement}.

\subsection{Contribution}
The tissue manipulation problem has already been approached by many researchers in \cite{alambeigi2019use,hirai2000indirect,shin2019autonomous,navarro2016automatic}, which have used either human-knowledge or machine-knowledge driven algorithms. In this paper, to the best of the authors' knowledge, for the first time, we propose an SL algorithm based on feature-based linear approximate Q-learning approach, which combines human knowledge with the learning algorithm to address this problem. In fact, we demonstrate that upon careful selection of simple yet effective features and reward, the developed linear approximate Q-learning algorithm is capable of accomplishing the desired manipulation task without any prior knowledge of tissue deformation behavior or camera calibration. 

\section{Methods}
\subsection{Problem Definition} This paper targets the task of indirect tissue manipulation with surgical robots that pinch \textit{Tissue Grasp Points (TGPs)}
to place \textit{Tissue Target Points (TTPs)} at \textit{Image Desired Points (IDPs)}, shown in Fig. \ref{SimEnvironment}b. The tissue phantom is manipulated at two TGPs representing the grasp locations of a surgical robotic system such as the Raven \cite{hannaford2013raven}. In operating rooms, TGPs, TTPs, and IDPs could be selected by the surgeon.

Q-learning is a popular Reinforcement Learning (RL) algorithm that attempts to find (near-) optimal policy by learning the $Q$ values of the underlying Markov Decision Process and taking an action that maximizes $Q$ at a given state. The most thorough form of Q-learning is to create a table of $Q$ values for each possible state-action pair which suffers from the curse of dimensionality problem \cite{sutton1998introduction}.
To remedy this problem, in our SL framework the $Q$ values are approximated with a linear function of features that are input based on human knowledge of the problem. 


\subsection{Tissue Manipulation with Approximate Q-learning}
As stated, the goal of the tissue manipulation task is to (indirectly) manipulate the TTPs to IDPs. Since the dynamics of the system (tissue) is not known \textit{a priori}, this problem can be viewed as an RL problem. Moreover, while camera calibration (intrinsic and extrinsic) information may be available for a surgical robotic system, it might not be accurate or may vary when the experimental parameters are changed. Hence in this paper we assume the robot has no camera calibration information to account for real robotics surgery scenarios. 
\subsubsection{Approximate Q-Learning}
To incorporate human knowledge into solving the RL problem as well as to avoid the curse of dimensionality, we deploy a linear function approximation approach \cite{sutton1998introduction,rivlin2003introduction} to approximate the Q function values given state and action ($Q^*(s,a) \approx Q(s,a;w)$):
\begin{equation}
    Q(s,a;w)=\sum_{i=1}^{N}w_if_i(s,a)
    \label{QForm}
\end{equation}
where N is the number of features, $w_i$ is the $i^{th}$ weight, $w$ is the vector of weights, and $f_{i}$ is the $i^{th}$ basis function. 

This method allows incorporating human knowledge by handcrafting basis functions (known as \textit{features}) to create the function approximator. 
Although features must be selected with domain specific knowledge, in real-world robotics applications learning algorithms with task-specific representations/knowledge are suggested \cite{gu2016continuous}. In fact, in this paper we show that by carefully selecting simple yet effective features, a linear function approximator is rich enough to capture and accomplish the very complicated task of robotic manipulation of deformable tissue. 

In linear approximate Q-learning, the weights, $w$ in (\ref{QForm}), are updated to reduce the expected value of error norm (usually 2-norm) between the unknown $Q^*(s,a)$ value and the value of the Q approximator \cite{pandey2010approximate,mnih2013playing}:
\begin{equation}
    L(w) = \mathbb{E}\Big[\Big(Q^*(s,a)-Q(s,a;w)\Big)^2\Big]
    \label{Loss}
\end{equation}
where
\begin{equation}
Q^*(s,a) = \mathbb{E}\Big[R(s,a,s')+\gamma\text{ }\underset{a'}{\max}\text{ }Q(s',a')\Big]
\end{equation}
and $s$, $a$, $R$, $\gamma$, $s'$ are the current state, action, reward, discount factor and next state respectively. Differentiating the loss function in (\ref{Loss}) with respect to $w$, we obtain the gradient ($\nabla$) as: 
\begin{equation}
    \nabla = \mathbb{E}\Big[\Big(R(s,a,s') + \gamma\text{ }\underset{a'}{\max}\text{ }Q(s',a')-Q(s,a)\Big)f(s,a)\Big]
    \label{OmegaUpdate}
\end{equation}
where $w$ is then updated based on gradient decent. In \eqref{OmegaUpdate}, $f(s,a)$ is the vector of features. 
\subsubsection{States and Actions}
The actual state of the problem contains the location of all of the nodes of the simulated tissue, which is 
intractable for computational purposes. Since the goal for tissue manipulation is to accurately move the TTPs to IDPs, we approximate the states of the system by only the x and y pixel coordinates of each of TTPs in image space (i.e. $s \in \mathbb{R}^4$).

Although actual surgical robot manipulators are capable of moving in three dimensions, in order to simplify the problem and enable the use of a single camera image, we constrained the motion of the TGPs to a plane parallel to and above the tissue surface. To reduce the action space dimension, an action set consisting of staying still, moving up, down, left, or right was selected for each TGP (i.e. $a \in \mathbb{R}^2$). This selection resulted in 5 possible actions for each TGP and, therefore, a total of 25 possible actions for both TGP points. Of note, all actions are defined in robot base frame as camera calibration information is assumed to be unavailable.

\subsubsection{Reward}
We deploy reciprocal Euclidean norm of the pixel error between IDPs and TTPs: 
\begin{equation}
    R(s,a,s') = R(s) = \frac{1}{\sqrt{e^Te + \epsilon}_s}
    \label{Reward}
\end{equation}
 where $e \in \mathbb{R}^{4}$ is the pixel error vector between IDPs and TTPs. Note that the pixel information of TTPs is obtained with an image processing thread and
$\epsilon_s$ is a small number ($10^{-8}$) to avoid reward divergence to infinity. 
\subsubsection{Features}
In order to satisfactorily build a good linear approximation function for $Q^{*}(s,a)$, it is necessary to select a set of features that can properly distinguish what is a good action in a given state for the task of tissue manipulation. The only simple human reasoning needed to select such features is that the actions from TGPs (instruments) changes the states of TTPs, which implies unknown coupling between them (see Fig. \ref{SimEnvironment}c). We embed these couplings into our Q-learning framework by selecting features in the form of x or y distances between each TTP and corresponding desired points, multiplied by all permutations of actions from TGPs (represented by numeric values). For the x actions of the $k^{th} \text{ } (k \in \{1,2\}$) TGP, $a_{kx}$ is set to -1 for an action in the negative x direction, 0 for no movement in the x direction, and +1 for an action in the positive x direction. Similarly, for the y actions of the $k^{th} \text{ } (k \in \{1,2\}$) TGP, $a_{ky}$ is set to -1 for an action in the negative y direction, 0 for no movement in the y direction, and +1 for action in the positive y direction. Therefore, a total number of 16 (4 couplings $\times$ 2 relative x or y pixel positions between TTPs and IDPs $\times$ 2 TTP-IDP pairs) are selected which can be expressed in the form:
\begin{equation}
f_i(s,a)=(d_{kl} - t_{kl})\text{ }a_{pq}, \text{       } i = 1,2,...,16
\end{equation}
where $t$ and $d$ represent TTP and IDP respectively, $k,p \in \{1,2\}$ indicating the $k^{th}$ or $p^{th}$ point, $q,l \in \{x,y\}$ indicating the coordinate or direction. 
These features work to consider how the action of each grasp point affect the x and y distances between each target point and corresponding desired point. However, none of these features will output the "no move" action. Intuitively, the TGPs should stop moving if the task is complete and the TTPs are very close to the IDPs. To enable this functionality, a $17^{th}$ feature was added that is essentially a Boolean statement to output a 1 if the reward is above a certain threshold and the "no move" action is selected. It can be formulated as:
\begin{align}
    f_{17}(s,a)= 
\begin{cases}
    1,       & \text{if }R(s,a=0,s')>0.08\\
    0,              & \text{otherwise}
\end{cases} \label{f17}
\end{align} 
The threshold selected was 0.08, which roughly corresponds to when the target points mostly overlap the desired points in the image. 
Increasing or decreasing the threshold therefore correspondingly changes the amount of training needed. 

\subsubsection{Episodic Training}
It is unlikely for random actions along a single trajectory to explore a wide variety of state-action pairs. Hence, a training episode structure was used where the state configuration is reintialized at the beginning of each episode with a total number of $N_{action}$ actions. This is repeated for a total number of $N_{episode}$ episodes. 

\subsubsection{Exploration Strategy}
To properly train a Q-learning algorithm, it is necessary to find a balance between exploring unknown state-action pairs and exploiting the current optimal policy. This is known as \textit{exploration-exploitation trade-off} in the literature \cite{kober2013reinforcement}. 
Some of the standard methods that were attempted for this task include $\epsilon$-first, $\epsilon$-greedy, and $\epsilon$-decreasing. 
The most success was an $\epsilon$-decreasing strategy according to \eqref{Epsilon}
\begin{equation}
    \epsilon=\max(0.1,1 - \frac{N_{action}\times n_{episode}+n_{action}}{N_{episode}\times N_{action}})
    \label{Epsilon}
\end{equation}
where $n_{action}$ is the number of actions that have been taken in the current episode, $n_{episode}$ is the current episode number from 0 to $N_{episode}-1$.

\subsubsection{Learning Rate and Discount Factor}
The learning rate, $\alpha$ is the step size in the gradient descent of updating weights in \eqref{OmegaUpdate}. There is a trade-off between smaller or larger values of learning rate. The larger $\alpha$ will enable faster convergence but has a potential for divergence while smaller $\alpha$ has higher chance for convergence.
A common strategy for learning rate is to set it to the following form for some constant $K$:
\begin{equation}
\frac{K}{K+N_{action} \times n_{episode}+n_{action}} 
\end{equation}
 In this paper, we deployed a cyclical learning rate (as proposed in \cite{SmithCyclic}) in the form mentioned above. Such a learning rate oscillates between a lower and upper bound to overcome saddle points and increase convergence speed. 
 This was chosen as it causes convergence of weights as the number of actions approaches infinity, but also periodically resets the learning rate to a higher value to overcome saddle points and correct for noise due to visual occlusion. Discount factor was set to $\gamma=0.9$.

\begin{algorithm}[!b]
\caption{Approximate Q-learning}\label{RLAl}
\begin{algorithmic}[1]
\State $\textit{set}$ $n_{episode}$ $\textit{to}$ $0$
\State $\textit{set}$ $N_{episode}$ $\textit{and}$ $N_{action}$ \textit{values}
\State \begin{varwidth}[t]{\linewidth}
$\textit{initialize weights to random numbers uniformly}$\par
$\textit{distributed between -0.5 and 0.5}$ \end{varwidth} \vspace{5pt}
\While {$(n_{episode}$ $<N_{episode})$}
\State $\textit{initialize configuration of simulation environment}$
\State $\textit{set}$ $n_{action}$ $\textit{to}$ $0$
\State $\textit{increase}$ $n_{episode}$ $\textit{by 1}$
\While{$(n_{action}$ $<N_{action})$}
\State $\textit{set }$ $\epsilon$ $\textit{and}$ $\alpha$
\State $\textit{select $\epsilon$-greedy action a}$
\State $\textit{calculate all}$ $f_i(s,a)$ \textit{and} $Q(s,a)$
\State $\textit{take action a}$
\State $\textit{calculate}$ $R(s,a,s')$ \textit{and} $\underset{a'}{max}\textit{ }Q(s',a')$
\State $\textit{update all}$ $w_i$ $\textit{according to \eqref{OmegaUpdate}}$
\State $\textit{increase } n_{action} \textit{ by 1}$
\EndWhile
\EndWhile
\label{alg:Algorithm}
\end{algorithmic}
\end{algorithm}

\subsubsection{Approximate Q-learning Algorithm}
The general procedure of the approximate Q-learning algorithm for training used in this paper is detailed in algorithm 1. After training, $\epsilon$ is set to zero, $w_i$ stops updating, and deterministic optimal trained policy is used for testing. 

 
 
 
 
 
  
  
  
  
  
  

\section{Simulation Study and Evaluation}
To evaluate the efficacy of the proposed SL algorithm based on linear approximate Q-learning for robotic tissue manipulation, we first developed a simulation environment. We then designed tissue manipulation simulation experiments based on real surgical scenarios and evaluated the performance of the algorithm from both learning and tissue manipulation point of views.

\subsection{Simulation Environment}
The simulation environment was created using CHAI3D library \cite{Conti03}. OpenCV library \cite{bradski2000opencv} was deployed for image processing purposes. Fig. \ref{SimEnvironment} depicts the simulation environment. 
The high frequency ($\sim$1 kHz) \textit{Dynamics Thread} solves the underlying partial differential equations (PDEs) in real time to update the position of each node (shown by blue spheres in Fig. \ref{SimEnvironment}a). The low frequency ($\sim$60 Hz) \textit{Graphics Thread}
receives the data from dynamics thread and renders the image by updating the image pixels. This rendering is internally taken care of by means of OpenGL library \cite{hearn2004computer}. For the purpose of this work, a new thread, \textit{Image Processing Thread}, was added that runs at ($\sim$60 Hz) and performs image filtering to obtain an estimate of the states for the problem.

\begin{figure*}[t!]
    \centering
    \includegraphics[scale=0.7, trim={6.7cm 0.2cm 6cm 0cm},clip]{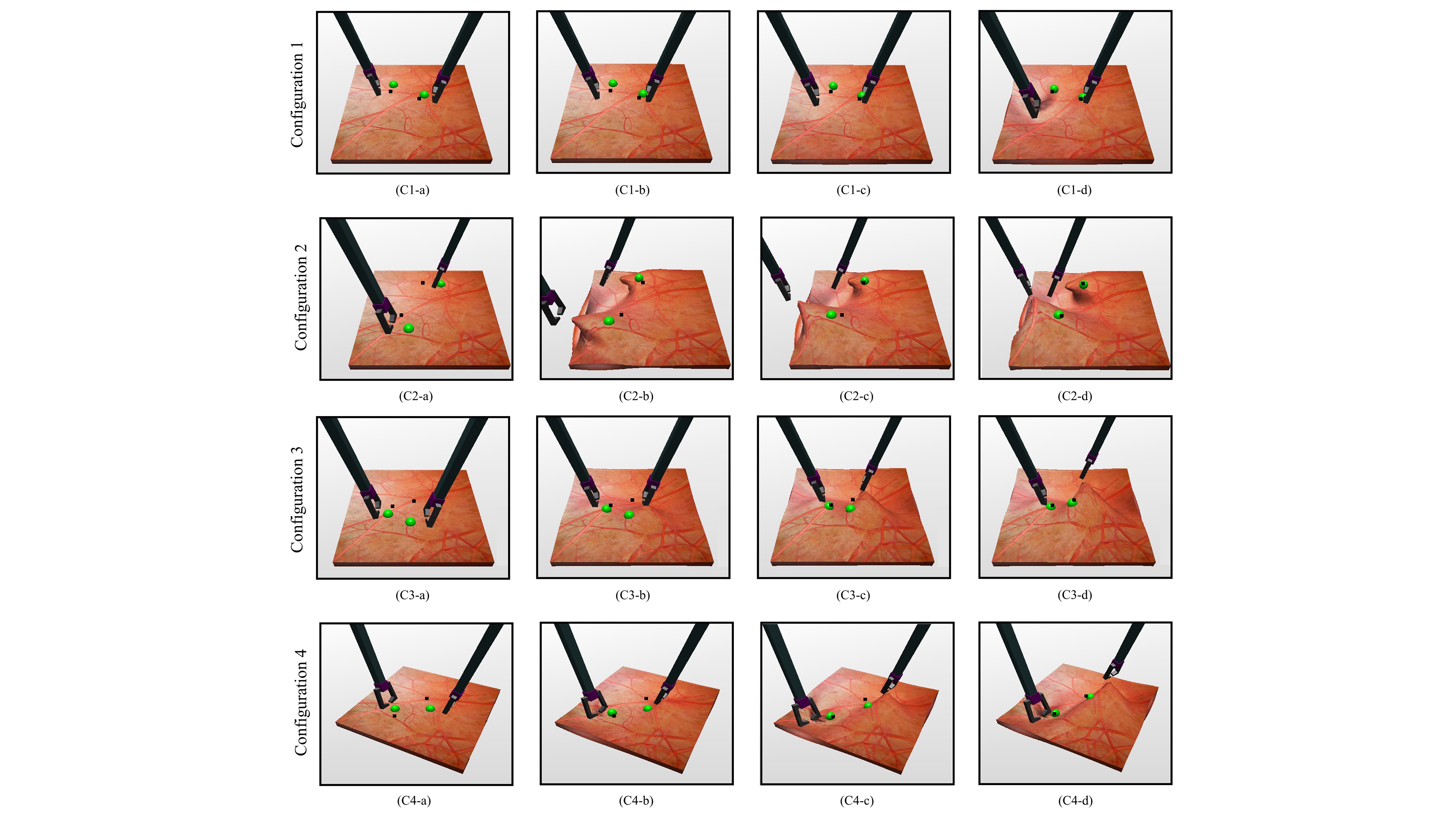}
    \caption{Example successful policies. Four configurations (C1-C4) are displayed for representative progressing actions during testing episodes from (a) \textit{initialization} to (d) \textit{task completion}. Each configuration represents a different set of initial TTPs, IDPs, and TGPs. Additionally, in (C4), the camera configuration is changed. In order to maintain simulation stability, it was necessary to manipulate TGPs via simulated springs with low stiffness connected to the graspers. As such, small gaps between graspers and the tissue are observable for large deformations.}
    \label{Multiple}
\end{figure*}
\subsubsection{Simulation Model} 
The GEL module of the open-source CHAI3D library \cite{Conti03} is used to simulate the motion/force of a real deformable tissue. The model consists of a rectangular prism structure of $25\times25\times1$ \textit{Nodes} (see Fig. \ref{SimEnvironment}b). The nodes of each lateral surface of the model are held fixed. Each node is connected to its adjacent nodes by means of \textit{Links} which specify how the motion of the nodes affect each other. The model takes into account the physical properties of each node (i.e. mass and radius) along with the physical properties of each link (i.e. linear and rotational stiffness and damping). This general structure is then used to calculate the interaction forces and dynamics of the tissue. For the purpose of this paper, additional software was developed so any node can be grasped and moved.

\subsubsection{Vision Algorithm}
Since in real surgical scenarios the IDPs are expressed in endoscopic images by surgeons, proper image processing should be used to extract TTPs pixel information from the images in real time. The image processing thread receives the frame buffer of the CHAI3D virtual camera at 60 Hz which could be replaced by an endoscopic frame buffer for software testing on an actual robotic surgery system. While more advanced algorithms (i.e. based on Convolutional Neural Nets \cite{ren2015faster}) could be used for tracking TTPs, in this work we use filters including Gaussian Blur and color-based HSV for image processing. The x and y center of the smallest circle containing all the identified TTP pixels are used to represent the position of the TTPs.

\subsubsection{Q-Learning Algorithm}
Once the Vision Algorithm obtains an estimate of states (position of the TTPs), it inputs the estimated states to the Q-learning algorithm (\textbf{Algorithm 1}) upon a query. State estimation is then used to calculate $f_i(s,a), Q(s,a), R(s,a,s')$ and $\underset{a'}{max}\textit{ }Q(s',a')$ and it takes $\epsilon$-greedy action $a$. 
\begin{figure*}[t!]
    \centering
    \includegraphics[scale=0.52, trim={0cm 3.6cm 0.5cm 0cm},clip]{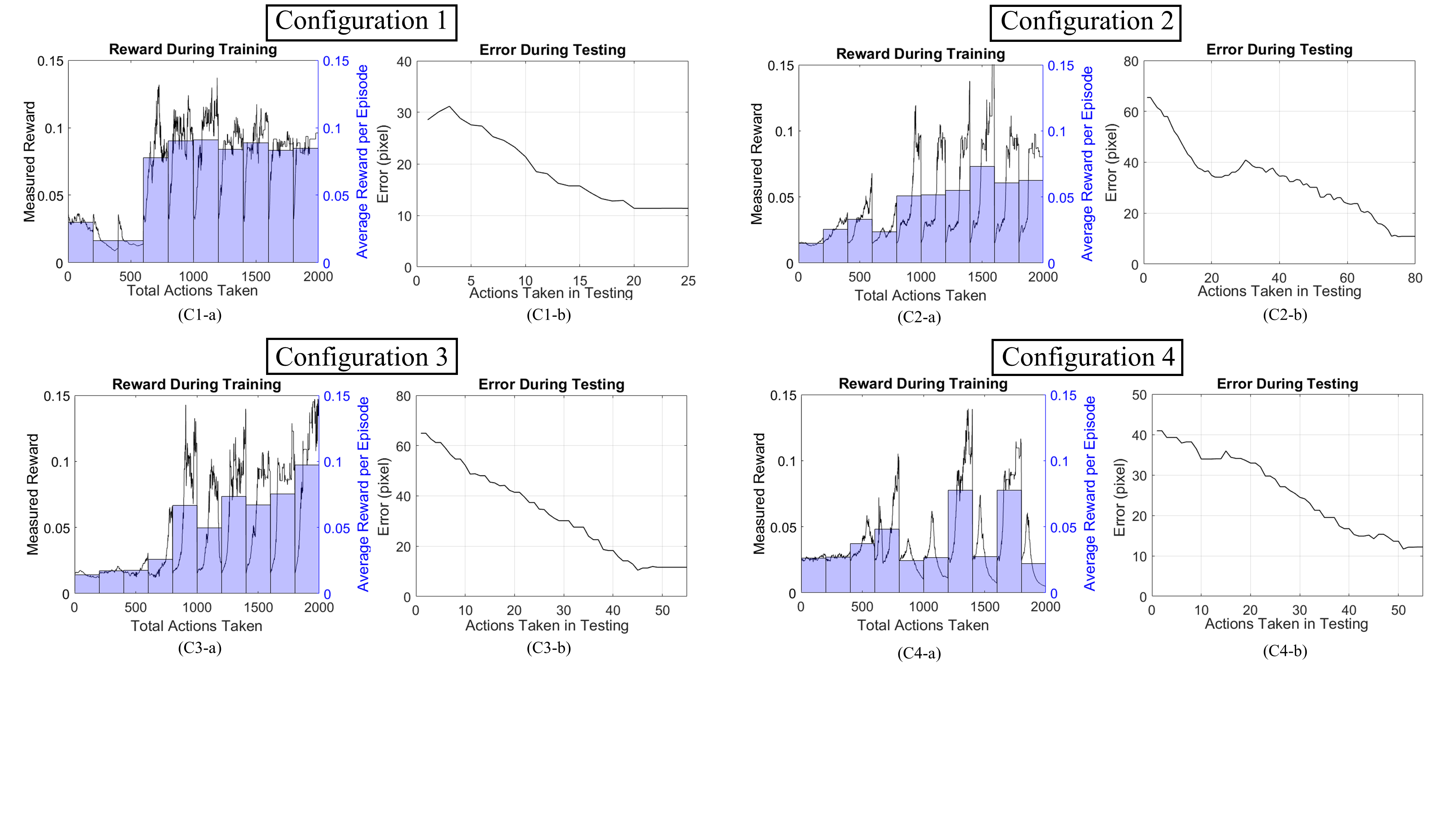}
    \caption{Reward during training and error during testing. (a) Rewards across all episodes of training for each configuration. The line plots indicate the reward per action while the bar plots display the average reward in each episode. (b) Euclidean error norm for single testing episodes for each configuration.}
    \label{RewardError}
\end{figure*}
\subsection{Simulation Experiment Design}
To design the experiments, we considered realistic situations that might happen during a robotics tissue manipulation procedure. For instance, locations of TGPs and IDPs are defined based on the clinical situations, surgical sites, and the preference of the surgeon. Therefore, the utilized manipulation algorithm needs to be robust to these variations. 
Moreover, although the camera calibration data might be available during real surgical robotic procedures, it might not be accurate or may vary when the experimental parameters are changed. Hence, it is highly desirable that the algorithm performance be independent of camera calibration and/or configuration. 
Considering these realistic cases, the performance of the algorithm was assessed on four different configurations, as shown is Fig. \ref{Multiple}. In the first three sets of experiments, the initial configurations of all TGPs, IDPs, and TTPs were changed (CASE I). For the fourth experiment, in addition to changing these points, the camera location was also altered (CASE II).

\subsection{Performance Evaluation Metrics}
To evaluate our algorithm, we assessed its performance from learning point of view as well as tissue manipulation perspective. In this regard, the learning performance was measured based on the reward during training episodes and error during testing. Tissue manipulation task was considered accomplished based on the Root Mean Square (RMS) error between TTPs and IDPs pixel coordinates.


\section{Results and Discussion}



In each training set, $N_{episode}=10$ episodes and $N_{action}=200$ actions were used.  Following the complete 2000 action training set, the trained policy was evaluated with testing episodes using a greedy policy ($\epsilon=0$) with $\alpha=0$.

Fig. \ref{Multiple} displays four time instances from (a) \textit{initialization} to (d) \textit{task completion} of robotic tissue manipulation for the described four different simulation studies. In these studies, the goal is to indirectly manipulate Green TTPs via TGPs to reach black IDPs. The first three rows are associated with the CASE I and the last row is associated with CASE II.

Fig. \ref{RewardError} shows the defined performance evaluation metrics including the reward during the training episodes of each simulation studies as well as the error which is reciprocal of the reward during testing, as defined in \eqref{Reward}. As can be observed in Fig. \ref{RewardError}, the algorithm correctly learns a policy that increases reward (on average) as training progresses. Of note, the displayed rewards are associated with all actions during the training sets  leading to the policies shown in Fig. \ref{Multiple}. 
For each configuration, reward increases both within episodes and across episodes. Note that the high frequency variation in reward at high values is due to the action size causing individual actions to significantly effect reward.

For testing episodes, Fig. \ref{RewardError} plots the Euclidean error norm between TTPs and IDPs against the number of actions taken. While configuration 1 does not require many actions, the learned policy directs the TGPs to quickly move the TTPs to the IDPs. Configuration 2 and 3 represent more challenging tasks that requires significant strain on the simulated tissue. Despite the increased complexity of the dynamics with large deformation, successful policies are learned. 
Configuration 4 further increases the complexity of the task by changing the configuration of the camera. 
As can be seen in Fig. \ref{Multiple}, instances C4-a through C4-d and Fig.\ref{RewardError}, instances C4-a through C4-b, the proposed algorithm was capable of learning weights that provide a successful policy. 

The converging errors in Fig. \ref{RewardError}, i.e. instances C1-b through C4-b, highlight the value of the defined feature in \eqref{f17}. This feature effectively serves as a completion criterion for the task by halting both graspers when the reward exceeds 0.08, which corresponds to a Euclidean error of 12.5 pixels, or roughly 8 pixel error for each TTP. It is worth noting that the four configurations shown represent very different initial configurations, yet the weight associated with the feature is properly trained in each case demonstrating the learning capabilities of the proposed SL algorithm. 

There were a number of challenges for both developing and running the proposed SL algorithm:  
   


\textbf{Visual Occlusion:}
Due to the simulation environment including avatars of both surgical robot graspers, as well as the use of an opaque tissue phantom model, it is not uncommon for vision of one or both TTPs to become occluded. Proper selection of TGPs relative to TTPs in the camera frame can help to prevent the graspers or dragged tissue from obscuring the TTPs during exploitation with a good policy. However, during exploration, the random actions will cause the grasper to pass over nearly all nearby points including the TTPs. When this occurs, it was found necessary to halt the update of all weights until proper vision is restored via actions selected by the current $\epsilon$-greedy policy. Additionally, if a grasper passes over a TTP, it is possible for the vision algorithm to incorrectly determine the center of the TTP, creating noise in the reward and features. 
The cyclical decreasing learning rate used in this work was found to assist in overcoming this noise. 

\textbf{Feature Selection:}
Feature selection is among most challenging part of feature-based Q-Learning as there is not a standard method for selecting features. Features must be selected based upon knowledge of the system and potentially subjective judgments for what equation should be optimized.

This work aimed to show that incorporation of simple features based on intuitive human knowledge were capable of solving the highly complicated task of simulated tissue manipulation with no camera calibration information with approximate Q-learning. The features chosen were selected to be the simplest and minimum number necessary to approach the task without knowledge of registration between camera, tissue, and robot. The developed algorithm converges to a successful solution, indicating that the simulation environment, image processing, and Q-learning all are designed reasonably. As such, the authors consider this work to be a successful implementation of a computational framework on a previously undocumented Q-learning problem. 

\bibliographystyle{ieeetr}
\bibliography{main}

\end{document}